\documentclass[10pt,twocolumn,letterpaper]{article}

\usepackage[pagenumbers]{wacv} 

\usepackage{graphicx}
\usepackage{amsmath}
\usepackage{amssymb}
\usepackage{booktabs}
\usepackage{multirow}
\usepackage{amsmath}
\usepackage{amssymb}
\usepackage{wrapfig}

\usepackage[pagebackref,breaklinks,colorlinks]{hyperref}

\usepackage[capitalize]{cleveref}
\crefname{section}{Sec.}{Secs.}
\Crefname{section}{Section}{Sections}
\Crefname{table}{Table}{Tables}
\crefname{table}{Tab.}{Tabs.}

\def\wacvPaperID{751} 
\def\confName{WACV}
\def\confYear{2025}

\begin{document}

\title{Learning to Generate Long-term Future Narrations Describing Activities of Daily Living}

\author{Ramanathan Rajendiran \\ \small Centre for Frontier AI Research, \\ \small Agency for Science, Technology and \\ \small Research (A*STAR), Singapore
\and
Debaditya Roy \\ \small Institute of High Performance Computing, \\ \small Agency for Science, Technology and \\ \small Research (A*STAR), Singapore
\and Basura Fernando \\ \small Centre for Frontier AI Research, \\ \small Agency for Science, Technology and \\ \small Research (A*STAR), Singapore}

\maketitle

\begin{abstract}
Anticipating future events is crucial for 
various application domains such as healthcare, smart home technology, and surveillance. 
Narrative event descriptions provide context-rich information, enhancing a system's future planning and decision-making capabilities.
We propose a novel task: \textit{long-term future narration generation}, which extends beyond traditional action anticipation by generating detailed narrations of future daily activities.
We introduce a visual-language model, ViNa, specifically designed to address this challenging task. 
ViNa integrates long-term videos and corresponding narrations to generate a sequence of future narrations that predict subsequent events and actions over extended time horizons. 
ViNa extends existing multimodal models that perform only short-term predictions or describe observed videos by generating long-term future narrations for a broader range of daily activities. 
We also present a novel downstream application that leverages the generated narrations called future video retrieval to help users improve planning for a task by visualizing the future.
We evaluate future narration generation on the largest egocentric dataset Ego4D. 

\end{abstract}

\section{Introduction}
\label{sec:intro}


Anticipating human actions is invaluable for proactive decision-making and tailored interventions in several areas. 
In video surveillance, it is useful to anticipate suspicious activities or abnormal behaviour that indicates criminal or dangerous intent. 
For individuals with special needs, smart-home agents can assist them based on anticipated events, such as automatically adjusting lighting, opening a refrigerator or other environmental settings. 
More importantly, in areas such as healthcare monitoring for elderly patients, it is critical to anticipate emergencies based on their movements and behaviours, allowing caregivers or medical professionals to intervene promptly.
In all the aforementioned applications, one common thread is the need to understand and anticipate human behaviour at a fine-grained level with a precision that demands detailed visual understanding.

Humans attain detailed understanding by storing their experience as a series of distinct events through a process called event segmentation. 
Event segmentation plays an important role in the recognition of intentions, episodic memory and consequently, the ability to imagine future events\cite{zacks2007event,swallow2018role}.
Event segmentation is automatically performed by humans and is driven by observable visual inputs more than the knowledge of the activity, context, or by an observer’s belief \cite{zacks2004using, zacks2009segmentation}.
Therefore, descriptions of events allow us to better understand and interpret human behaviours in videos.


Anticipating the short and long-term future of humans from videos has garnered significant attention, particularly with the emergence of expansive egocentric datasets like EPIC-KITCHENS \cite{damen2022rescaling} and EGO4D \cite{grauman2022ego4d}. These datasets have provided unparalleled resources for studying human activities in their natural environments, driving research into anticipatory models that aim to forecast forthcoming actions and events. 
Anticipation, within this context, predominantly focuses on actions characterized by noun-verb combinations, such as ``cut wood'' and ``open fridge''. 
These formulations typically encapsulate the primary object involved in the event as the noun and the imperative verb denoting the action. 
However, verb-noun pairs often offer a limited perspective of the event without providing additional contextual details. 
\begin{figure*}[t!]
    \centering
\includegraphics[width=\linewidth]{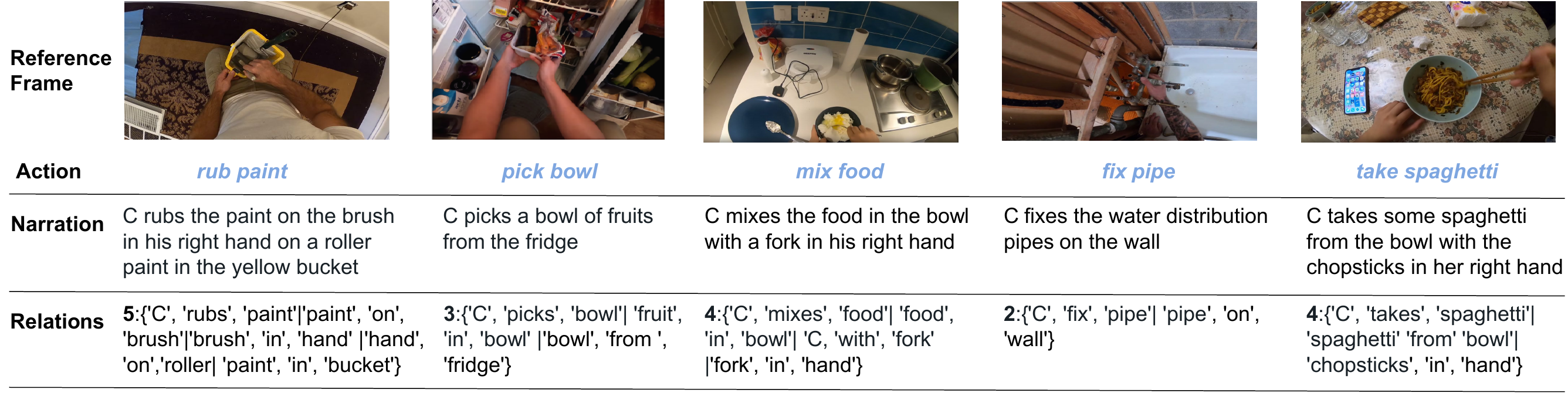}
    \caption{Narrations offer a detailed account of events, capturing multiple interactions between individuals and objects. Unlike verb-noun depictions of actions, narrations deliver a detailed perspective of an event within a video. By focusing on generating future narratives instead of mere actions, we gain a richer understanding of what's to come. In this context, C represents the person wearing the camera.}
    \label{fig:motivation}
\end{figure*}
%
In contrast, narrations such as ``cleans his hands on a fabric'' and ``puts the jerrycan on top of the table'' are \textit{event descriptors} that provide context around the actions, making it easier for the listener or reader to understand the circumstances, reasons, and potential outcomes associated with the actions. 
This context can include where, when, and how an action will take place, which is often as important as the action itself.
Beyond the simple noun-verb constructs, narrations encompass a broader spectrum of information, including the tools utilized to perform the action and the specific location where the action transpires. 
As shown in \cref{fig:motivation}, narrations cover one or more human-object and object-object relations.
Therefore, we propose a new task -- generation of long-term future narrations which closely resemble the human ability to imagine future events.

We propose to generate future narrations (e.g. 48 seconds into the future) based on both observed videos and their narrations for three reasons. 
Firstly, learning from narrations alone can lead to a narrow understanding, missing non-verbal cues and context provided by videos of daily activities. Secondly, videos offer complementary visual information about the environment that is lacking in narrations. Finally, solely focusing on narrations can cause an over-reliance on verbal information. 
Our future narration generation task is challenging to the models as 
they have to learn the continuity in daily activities over longer time horizons.
Current large video-language models are trained to generate short answers \cite{yang2022zero} or accept only short sequences \cite{alayrac2022flamingo, pramanick2023egovlpv2}.
While models for dense-video captioning \cite{yang2023vid2seq,zhao2023learning,zhang-etal-2023-video} or automatic audio description \cite{han2023autoad,han2023autoadii} generate detailed captions, these methods describe the observed video content and not what will happen in the future.
Future step generation methods \cite{abdelsalam2023gepsan, Patel_2023_ICCV} are designed for tasks that have only 3-4 steps. 
Our task of long-term narration forecasting requires that we generate 20+ future narrations with no supporting future video.
To demonstrate future narration forecasting tasks, we develop a novel video-language model (ViNa) that accepts both videos of 48 seconds and more than 20 narrations as inputs and generates 20 or more future narrations.

\textbf{Contributions} (1) We propose a new challenging task called long-term future narration generation in daily activities. To solve this task, we develop a visual-language model called \textbf{ViNa}. We compare ViNa with the existing visual-language models. 
ViNa is adapted to predict the different granularities of human behaviours, \ie goal, step, substep, narrations, and actions. Furthermore, ViNa performs close-ended prediction as well as open-ended generation. 
(2) We demonstrate a novel downstream application called \emph{future narrations-based future video retrieval} that leverages the generated narrations. The retrieved future videos help users to improve planning for a task by visualizing the future steps.

\section{Related Work}
\label{sec:related-work}
\textbf{Generating Future in Procedural Videos.}
Approaches such as \cite{sener2019zero, bansal2022my, zhong2023learning} generate natural language descriptions for future key steps in procedural videos. 
Procedural or instructional videos  only capture steps that lead to the completion of a task such as cooking, assembly of household items, etc~\cite{song2023ego4d}.
In~\cite{zhong2023learning}, authors treat future step prediction as a matching problem where features of a future video clip are drawn from a Gaussian distribution and refined using a diffusion model.
The text embedding with the closest match to the refined future video clip feature is predicted as the next step in the sequence.
In \cite{sener2019zero, abdelsalam2023gepsan} future steps are generated in natural language from past steps using a text-based pre-trained encoder-decoder network.
In \cite{sener2019zero}, a text-based pre-trained encoder-decoder network is fine-tuned with a video encoder and frozen decoder to achieve observed video to future steps generation.
In \cite{abdelsalam2023gepsan}, video representation is used without finetuning during training which is termed as zero-shot modality transfer. 
In contrast, we generate future narrations of daily activities over a longer time horizon where unconstrained daily life actions need not be performed in any predefined order as in unscripted datasets such as EPIC-KITCHENS 100 \cite{damen2022rescaling} and EGO4D \cite{grauman2022ego4d}.
As there are thousands of unconstrained actions, correctly predicting a sequence of actions is fraught with uncertainty. 

\noindent
\textbf{Joint Video and Text Modeling}
has become popular with tasks such as video question answering and video captioning and recent successes of video-language models.
The success of video-language models relies on cross-modal supervision via aligned multimodal data used for pre-training with minimal finetuning on downstream tasks \cite{ashutosh2023hiervl, lin2022egocentric, cheng2023vindlu, zhao2023learning}.
Each modality --video/image and text-- is encoded using its own specific encoder. 
The encoders are then trained using contrastive learning to produce aligned features.  
There are another set of approaches that jointly encode frame-wise video tokens and text tokens using a single multimodal transformer encoder such as \cite{ seo2022end, yang2022zero, alayrac2022flamingo,luo2020univl, li2023lavender}.
Recently, \cite{pramanick2023egovlpv2} propose cross-modal fusion of visual and text encoders using gating. 
Another task which uses joint video and text modeling is multimodal video captioning \cite{seo2022end, zhang2024mm} where audio transcriptions are available along with videos.
The daily activity videos that we use in narration forecasting typically do not contain audio and hence we cannot extract transcriptions.
Furthermore, narrations are more reliable than transcriptions as they describe visual events but transcriptions can describe visual and non-visual events.
Narrations have been mainly used for pre-training or data augmentation as temporally aligned features for downstream tasks such as multiple-choice question-answering, video summarization, text-to-video retrieval, natural language query grounding, spatio-temporal grounding and localization ~\cite{pramanick2023egovlpv2, ramakrishnan2023naq, zhao2023learning, chen2024and, shen2024learning}. We propose an open-ended generation task that explores the potential of narrations for understanding and anticipating events in videos even without explicit temporal video-text alignment pretraining.
\section{Method}
\label{sec:method}
Now, we present our approach for generating future narrations.
First in \cref{sec:prob_statement}, we describe the problem formulation. 
We then present our ViNa model in \cref{sec:model_design}.

\subsection{Problem Formulation}\label{sec:prob_statement}
The observed video contains multiple events and each event is described in natural language by a corresponding narration. 
For example, the narration \textit{Camera wearer scoops soil from a sack into the flower pot with a hand trowel in his right hand} defines an event in a gardening video. 
Given the observed video from which $T$ frame representations are extracted $\{V_1, \cdots, V_T\}$ and the corresponding set of narrations describing the events are tokenized as $\{L_1, \cdots, L_N\}$, our goal is to learn a model that generates tokens of narrations of future events $\{\hat{L}_{N+1}, \cdots, \hat{L}_{N+K}\}$ as follows:
\begin{equation}
\label{eq:objective}
      \hat{L}_{N+1}, \cdots, \hat{L}_{N+K} = f_\theta(V_1, \cdots, V_T, L_1, \cdots, L_N) 
\end{equation}
where $\theta$ refers to the model's trainable parameters. 
Inspired by recent visual-language models \cite{yang2022zero}, we train an auto-regressive decoder model to generate future text tokens based on observed video frames and associated text tokens. 

\subsection{Future Narration Generation Model - ViNa} \label{sec:model_design}
\begin{figure}
    \centering
    \includegraphics[width=\linewidth]{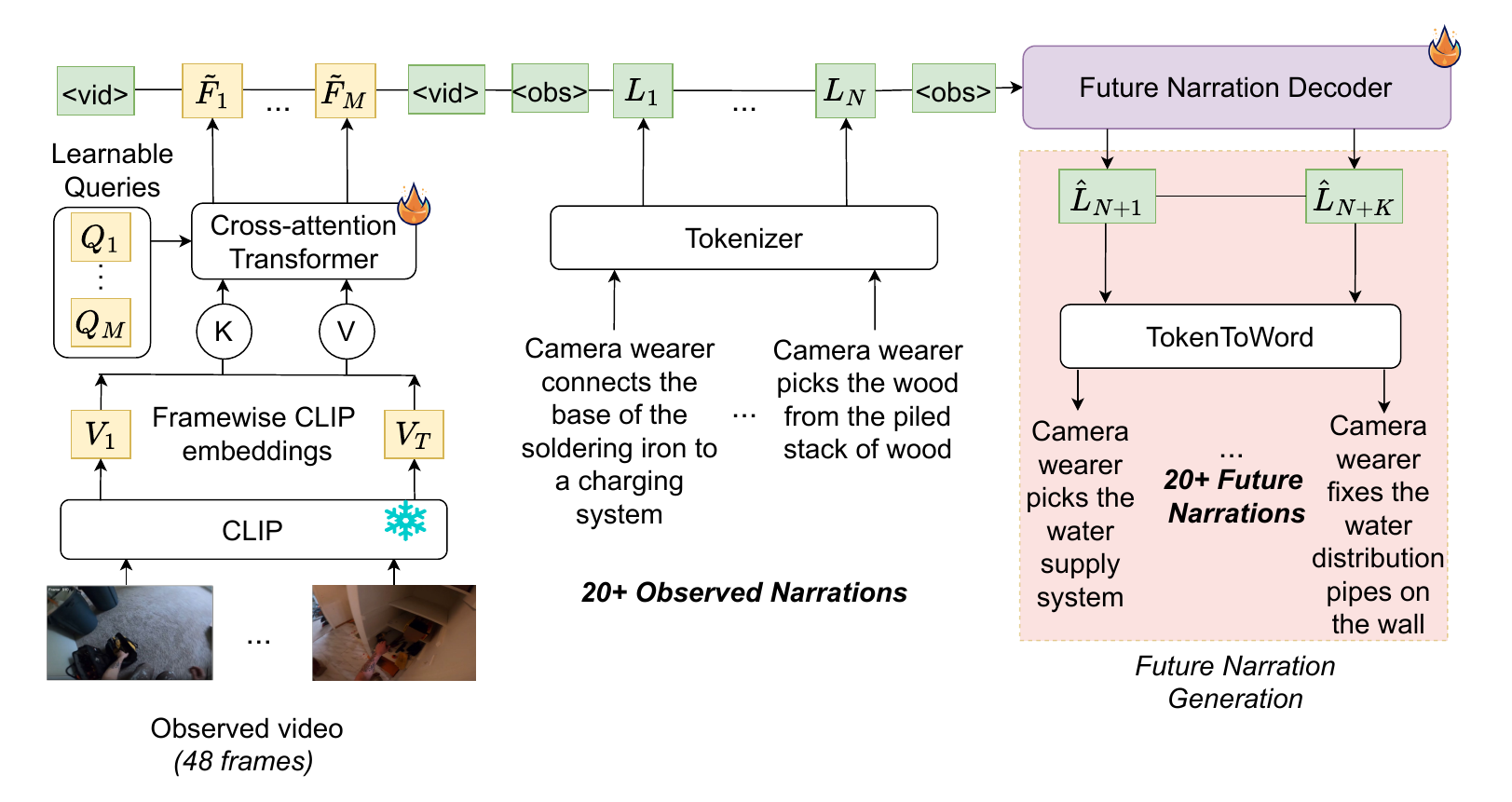}
    \caption{Proposed model ViNa for video-conditioned future narration generation. A frozen CLIP image encoder is used to obtain a frame representation of the observed video frames. Learned queries attend to the frame representation to obtain learned video tokens that summarize the events in the video. A decoder such as GPT-2 takes the learned video tokens and observed narration language tokens to generate 20+ future narrations.}
    \label{fig:model_fnp}
\end{figure}

The challenges in adapting existing vision-language models to our setting are multi-fold. 
First, summarizing the spatio-temporal information in long videos is quite challenging \cite{wu2022memvit}. 
Second, the narrations and video tokens are not equal in number as each event takes different duration which is known apriori. Therefore, it is challenging to align video and narrations pairwise \cite{pramanick2023egovlpv2}. 
Third, our model has to learn present context from both the narrations and video and how the context can generate diverse future narrations while still maintaining coherence.
We discuss how we design our proposed model to handle these challenges as outlined in~\cref{fig:model_fnp}.



To manage a long observed video\footnote{48 seconds cover an average of 20 narrations describing events.}, we sample frames and compute a spatial encoding of each frame $\{V_1, \cdots, V_T\}$ using CLIP image encoder \cite{radford2021learning}. We are inspired by dense video captioning of long videos \cite{yang2023vid2seq}
and keep the CLIP backbone frozen for efficiency.
We sample the observed video at a particular frame rate which can lead to our sampled frames containing more information about longer events compared to shorter events.
However, the observed narrations describe both long and short events in single sentences. 
Therefore, we propose to learn latent representations from the videos that represent different events in the observed video.
We design a transformer network inspired by Q-Former \cite{li2023blip} using a set of latent learned queries $Q = W_Q\{Q_1, \cdots, Q_M\}$ and the frame tokens as keys and values $K = V = W_{KV} \{V_1, \cdots, V_T\}$. 
Here $W_Q, W_{KV}$ are learned parameters.
A cross-attention Transformer~\cite{vaswani2017attention} then produces latent representations which we call learned video tokens denoted by $\tilde{F}_1, \cdots, \tilde{F}_M = softmax (QK^\top) V$ --see~\cref{fig:model_fnp}.
The aim of the cross-attention transformer in our setting is to extract learned video tokens that capture different events in the observed video. 
We learn video tokens when fine-tuning the language model on the task of future narration generation whereas Q-Former~\cite{li2023blip} learns latent queries during pre-training with image-text contrastive learning followed by using the learned latent queries for fine-tuning on downstream tasks. 

After obtaining the video representation, we formulate a multimodal input that combines the learned video tokens and the observed narration tokens.
Inspired by Vid2seq \cite{yang2023vid2seq} and ClipCap \cite{mokady2021clipcap}, we treat the learned video tokens as prompt vectors that adapt to the text generation model. 
We create a long input token sequence containing the learned video and observed narration tokens  $O = \{\tilde{F}_1, \cdots, \tilde{F}_M, \texttt{<vid>}, \texttt{<obs>}, L_1, \cdots, L_N, \texttt{<obs>}\}$ with separator tokens $\texttt{<vid>}$ and $\texttt{<obs>}$ representing the start and end of video and observed narration\footnote{We also tried learnable separator tokens but fixed separator tokens gave better performance.}.
To predict the future narrations, we propose to use a language decoder such as GPT-2 \cite{radford2019language} following image and video description models \cite{mokady2021clipcap, zhao2023learning}.
These decoder models are not trained for generating future narrations of daily activities and we need to train the decoder.
Decoder models are trained to generate an output sequence which is shifted right by one token from the input sequence. 
Thus, at training, we supply the future narrations along with the observed video and narrations to the decoder $O_{train} = \{O, \texttt{<fut>}, L_{N+1}, \cdots, L_{N+K}, \texttt{<fut>}\}$. 
During testing, we only provide the learned video and observed narration token sequence $O$.
We show our future narration generation model ViNa in \cref{fig:model_fnp}.

\section{Experimental Results}
\label{sec:experiments}
The experimental section is organized as follows: first, we describe the dataset curation methodology and the evaluation metrics for future narration generation and the downstream tasks in~\cref{sec:dataset_curation}. The implementation setting is defined in~\cref{sec:setting}. We then explore the choice of the future narration decoder model in~\cref{sec:future_narration_generation_arch}.
Next, we explore how existing vision-language models can be adopted for future narration generation and compare our method in~\cref{sec:future_narration_generation_performance} to demonstrate the effectiveness of our proposed architecture.  We then shift focus to discuss an application of how the generated narrations can be used for future video clip retrieval in~\cref{sec:future_video_retrieval}.
In~\cref{sec:ablation}, we perform ablation studies on different video representations, the effect of the number of learned video tokens and the length of the decoded future narrations. We also investigate the effect of predicted observed narrations instead of ground-truth narrations. Finally, we present qualitative results in~\cref{sec:qualitative_results}.

\subsection{Dataset Curation and Evaluation Metrics}
\label{sec:dataset_curation}

We build our long-term narration generation benchmark on the Ego4D dataset. 
Ego4D is the largest first-person video dataset sourced from humans around the world going about their daily life activities. It contains more than 9,600 manually annotated videos spanning close to 3,900 hours. At approximately 13.2 sentences for each minute of video, 3.85 million narrations describe precise fine-grained events of humans wearing a camera. The narrations are diverse with 1,772 unique verbs and 4,336 unique nouns. Narrations essentially encode the complexity of relations between nouns. Approximately, 50.83\% of the narrations contain multiple relationships involving 2 or more nouns while actions capture only a singleton relation.

In the Ego4D~\cite{grauman2022ego4d} dataset, more than 50\% of the narrations encode relationships involving 2 or more nouns (person, object, place, etc.). On the other hand, majority of the narrations in Epic-Kitchens 100~\cite{damen2022rescaling} are verb-noun combinations, only 9.28\% of the  narrations contain multiple relationships. Therefore, we use the Ego4D dataset due to its rich and diverse narrations.

Existing forecasting tasks in Ego4D are tailored to short-term hand object prediction and long-term activity prediction. However, these involve predicting simplistic objects and actions while our future narration generation task involves predicting more informative unstructured text. Since, the new task requires open-ended generation over a longer period, it is essential to curate a new benchmark with a large number of training examples involving observed videos, observed narrations and future narrations.

\textbf{Future Narration Generation Task.} We now describe our future narration generation task.
We focus on the subset of Ego4Dv2 videos that are part of the ``Episodic Memory'' and ``Forecasting Hands-Objects'' subsets. 
As we want to generate narrations over the long-term, we select videos which are at least 96s in length where we observe the first 48s and generate narrations for the next 48s.
This choice is also made to keep the number of narrations comparable with the Long-term Action Anticipation task where we need to predict 20 actions in the future which approximately corresponds to 48s. 
We divide the videos which fit the aforementioned criteria into train, validation and test splits where the videos in one split are not shared with the others.
Each video is then divided into chunks of 8 minutes to handle the scale. From each 8 minute chunk, we first obtain five uniformly sampled pairs of contiguous 48s observed and 48s future windows with starting points {0, 96, 192, 288, 384}.
For the training set, we augment the dataset with another five samples per chunk with random starting points within the chunk. We finally have 60,851 training, 3,702 validation and 4,077 test samples.
Each observed or future sample has approximately 20 narrations.
We evaluate the performance of our future narration generation models on standard captioning metrics - Rouge-L~\cite{lin2004rouge}, BLEU-4~\cite{papineni2002bleu}, METEOR~\cite{banerjee2005meteor}, CIDEr~\cite{vedantam2015cider} and the BertScore~\cite{zhang2019bertscore} similarity measure.

\subsection{Implementation Details}
\label{sec:setting}
For obtaining the visual features, we uniformly sample one frame per second from the observed video. Visual features are extracted using the pre-trained CLIP ViT-L/14 model, while narration features are acquired using the language model's token embedder. We perform fine-tuning on the pretrained 12-layer GPT-2 language model for all experiments conducted. We train our the AdamW optimizer with a cosine-decayed learning rate schedule for 3 epochs. The default learning rate is $5e^{-5}$. All the experiments were run on an Nvidia A40 GPU with 46 GB of memory. Our framework uses the frozen visual features and requires considerably less training time (approximately 4 hours) compared to other vision-language models which require extensive pretraining. We will release our code and the annotations.

\subsection{Future Narration Generation Performance}
\label{sec:future_narration_generation_arch}

We first investigate the choice of the GPT-2 decoder for future narration generation vis-a-vis other language models in \cref{tab:arch-design-gt}. In ViNa-LLama-13b, we replace the tokenizer for the observed narrations with an OpenLLama-13b tokenizer and the GPT-2 decoder with an OpenLLama-13b decoder. Similarly in ViNa-LLama3-8b, we use the BPE tokenizer and the Llama3-8b decoder.  We trained the LLama models with parameter efficient fine-tuning - QLoRa~\cite{dettmers2024qlora}. Finally, in ViNa-T5, we replace the tokenizer for observed narrations in ViNa with a T5 encoder \cite{raffel2020exploring} and the GPT-2 decoder in our model with the T5 decoder inspired by \cite{yang2023vid2seq}. We fine-tune the T5 encoder and decoder.

\begin{table}[ht]
\scriptsize
\centering
\begin{tabular}{l|l|c|c|c|c|c} 
\hline
& \textbf{Model} & \textbf{R-L} &  \textbf{B-4} &  \textbf{M} &  \textbf{C} &  \textbf{BERT} \\ \hline
\multirow{4}{*}{Val} & ViNa-Llama3-8b & 31.73 & \textbf{26.76} & \textbf{32.66} & 10.92 & 22.92 \\
& ViNa-Llama-13b & 32.51 & 24.22  & 31.11 & 13.40 & 27.51 \\
& ViNa-T5 & \textbf{33.88} & 19.39 & 28.92 & \textbf{21.49}  & \textbf{32.69} \\
& ViNa-GPT2 & 31.85 & 26.35 & 31.98 & 11.83 & 26.83 \\
\hline
\hline
\multirow{4}{*}{Test} & ViNa-Llama3-8b & 32.10 & 25.57 & \textbf{33.70} & 12.78 & 23.87 \\
& ViNa-Llama-13b & 32.98 & 25.31 & 32.17 & 16.68 & 28.21 \\
& ViNa-T5 & \textbf{34.34} & 20.35 & 29.84 & \textbf{25.52} & \textbf{33.51}  \\
& ViNa-GPT2 & 32.42 & \textbf{26.81} & 33.10 & 11.81 & 27.57 \\
\hline
\end{tabular}

\caption{Comparing future narration generation performance of ViNa decoders on ground-truth narrations. Rouge-L (R-L), Bleu-4 (B-4), Meteor (M), Cider(C). ViNa-T5 outperforms other models on Rouge-L, CIDEr and BERT-Score metrics while ViNa-GPT2 and ViNa-LLama3-8b improves on BLEU-4 and METEOR. 
}
\label{tab:arch-design-gt}
\end{table}
As seen in \cref{tab:arch-design-gt}, the larger language models -- ViNa-Llama-13b and ViNa-Llama3-8b, perform comparable to GPT2 and T5 models but do not surpass the performance on 4 out of 5 metrics. 
We attribute this to the lack of data for effective full fine-tuning of such large models on a new task. 
On the other hand, smaller models such as T5 and GPT-2 that can be trained effectively for the future narration generation task.
T5 being the smallest model with 220M performs the best on 3 out of 5 metrics while the next largest GPT2 with 1.5B performs the best on BLEU-4.

\subsection{Comparison with Visual-Language Models}
\label{sec:future_narration_generation_performance}
As there is no prior work addressing the task of long-term future narration generation conditioned on videos, we compare our method against existing methods which can be adapted for our task.
(1) CLIPCap~\cite{mokady2021clipcap}: We map the spatial CLIP features of each frame in $\{V_1, \cdots, V_T\}$ to a learned frame prefix of prefix-length $P$.
The frame-wise learned prefixes are then concatenated to obtain $\{\tilde{V}_{11}, \cdots, \tilde{V}_{1P}, \tilde{V}_{21}, \cdots, \tilde{V}_{2P},\cdots, \tilde{V}_{T1}, \cdots, \tilde{V}_{TP}\}$. We then combine the concatenated frame-prefixes with the observed narration tokens to create the observed sequence as before. These tokens act as prompt vectors to the text generation model.
(2) Gated Cross-Attention: An alternate strategy to visually condition text-decoders is via introducing additional layers into the transformer layers of the language model which cross-attend to the visual CLIP features. This is termed as gated cross-attention in ~\cite{alayrac2022flamingo}. 
(3) Vid2Seq \cite{yang2023vid2seq}: performs dense video captioning with the video and transcribed speech as input. We change the transcribed speech input with observed narration. Vid2Seq encodes observed narrations with a T5 encoder \cite{raffel2020exploring} and combines the output with observed video tokens. It then uses a T5 decoder to generate observed video captions which we retrain to generate future narrations.

\begin{table}[ht]
\scriptsize
\centering
\begin{tabular}{l|l|c|c|c|c|c} 
\hline
& \textbf{Model} & \textbf{R-L} &  \textbf{B-4} &  \textbf{M} &  \textbf{C} &  \textbf{BERT} \\ \hline
\multirow{5}{*}{Val} & CLIPCap*~\cite{mokady2021clipcap} & 31.68 & 25.95 & 32.02 & 10.60 & 26.51\\
& Gated Cross-Attention*~\cite{alayrac2022flamingo} & 31.47 & 25.68 & 31.60 & 10.87 & 26.17 \\
& Vid2Seq*~\cite{yang2023vid2seq} & 33.78 & 19.63 & 29.14 & 21.35 & 32.45 \\ 
\cline{2-7}
& ViNa-T5 & \textbf{33.88} & 19.39 & 28.92 & \textbf{21.49}  & \textbf{32.69} \\
& ViNa-GPT2 & 31.85 & \textbf{26.35} & \textbf{31.98} & 11.83 & 26.83 \\
\hline
\hline
\multirow{5}{*}{Test} & CLIPCap*~\cite{mokady2021clipcap} & 32.14 & 26.09 &	33.06 & 11.93 & 27.20 \\
& Gated Cross-Attn*~\cite{alayrac2022flamingo} & 31.75 & 25.76 & 32.55 & 11.22 & 26.55 \\
& Vid2Seq*~\cite{yang2023vid2seq} & 34.26 & 20.59 & 30.09 & 23.47 & 33.15 \\
\cline{2-7}
& ViNa-T5 & \textbf{34.34} & 20.35 & 29.84 & \textbf{25.52} & \textbf{33.51}  \\
& ViNa-GPT2 & 32.42 & \textbf{26.81} & \textbf{33.10} & 11.81 & 27.57 \\
\hline
\end{tabular}
\caption{Comparing ViNa with existing visual-language models on future narration generation. ViNa-T5 outperforms other models on Rouge-L, CIDEr and BERT-Score metrics while ViNa-GPT2 is the best on BLEU-4 and METEOR. *Our implementation of existing video-language models. 
}
\label{tab:sota}
\end{table}

As shown in \cref{tab:sota},
ViNa-GPT2 outperforms CLIPCap even when both use the same GPT-2 decoder. 
This is due to ViNa-GPT2 learning a video representation as a prefix while CLIPCap learns an image prefix.
ViNa-GPT2 also outperforms other video based approaches such as Vid2Seq and gated cross-attention on BLEU-4 and METEOR.
ViNa-T5 outperforms Vid2Seq on Rouge-L, CIDEr and BERT-Score metrics while sharing the same T5 encoder-decoder backbone. 
Overall, ViNa-T5 is the most efficient in terms of parameters (220M) and effective on most generation metrics.

\subsection{Narrations for Future Video Retrieval}
\label{sec:future_video_retrieval}
We show an application of future narration generation called \textit{retrieval of future videos}.
This application can help users improve planning by visualizing the future steps in a task. 
Text-to-video generation models still struggle to generate realistic long-term videos from text prompts \cite{singer2022make}. 
Instead, it is useful if we can retrieve future videos from the generated narrations.
We divide every future video into a sequence of clips each described by a narration. 
For every video clip (sampled at 16 fps), we obtain its embedding using the EgoVLPv2 video encoder that is trained on EGO4D narrations and videos \cite{pramanick2023egovlpv2}. We end with a video embedding sequence for every future video $E_V = \{E_{V1},\cdots,E_{VM}\}$.
To retrieve the videos, we need to compare it to the generated narrations.
\begin{table}[ht]
\centering
\scriptsize
\begin{tabular}{l|l|l|l}
\hline
\textbf{Narration} & \textbf{R@50} & \textbf{R@75} & \textbf{R@100} \\ \hline
Random & 1.07 & 1.83 & 2.13 \\ \hline
GT Narration & 17.42 & 22.03 & 25.71 \\ \hline
\begin{tabular}[c]{@{}l@{}} Gen Narration - \\Video@1fps\end{tabular} & 1.32 & 2.01  & 2.67  \\ \hline
\begin{tabular}[c]{@{}l@{}} Gen Narration - \\ Video@16fps\end{tabular}& \textbf{17.04} & \textbf{20.57} & \textbf{23.22} \\ \hline
\end{tabular}
\caption{Performance on Future Video Retrieval. ViNa generated narrations almost reach ground truth narration performance.}
\label{tab:future_video_retrieval}
\end{table}

In the future narration generation setup, we generate a sequence of 20+ future narrations for every observed video.
We use the EgoVLPv2 text encoder to obtain an embedding for each narration. This gives a narration embedding sequence $E_T = \{E_{T1},\cdots,E_{TK}\}$. 
We compute the dot product $<E_V,E_T>$ to measure the similarity between every pair of narration and video embedding sequence in our test set.
Given that we have 4,077 test samples, we use Recall@$k$ where $k=50,75,100$, to measure retrieval accuracy that means the ground truth is in the first $k$ choices based on similarity.
\cref{tab:future_video_retrieval} shows the performance of future video retrieval.
Ground truth narrations are matched to the EgoVLPv2 video sequence embedding sequence.
We show that the generated narrations almost reach ground-truth narration performance.
We consider another baseline that is sparse sampling of videos at 1 fps that can improve the speed of embedding computation and retrieval. We extract CLIP image embeddings of the sampled images and compute the similarity with the CLIP text embeddings of the narrations.
The sparse video clips while being faster to compute perform similar to random baseline.
Therefore, we conclude that events require dense sampling for effective retrieval and generated narrations are as effective as ground truth narrations.

\subsection{Ablations}
\label{sec:ablation}
We perform various ablations on the video encoder, the length of the decoder tokens and the number of the learned video tokens in our model. We also study the importance of observed videos and using predicted narrations. 
\subsubsection{Comparison of Different Video Representations}
In \cref{tab:vis-encoder-ablations}, we compare different video representations for future narration generation. All these representations are obtained by changing the video representation module in ViNa-GPT2. First, we directly combine CLIP and EgoVLPv2 embeddings of observed video frames with the observed narration tokens. The baseline CLIP features perform better than EgoVLPv2 features although EgoVLPv2~\cite{pramanick2023egovlpv2} is pre-trained on the Ego4D dataset. Hence, we show that pre-training is not necessary for future narration generation. 
We also compare the performance of learning video tokens from CLIP features through a transformer with 12 self-attention layers. Future generation performance does not improve with learned video tokens using a transformer. 
We also investigate the performance of learned video representation when they are combined with the CLIP features and learned using a self-attention transformer with 12 layers, inspired by \cite{mokady2021clipcap}. 
We compare all the video representations to our approach of obtaining video representation from learned queries using cross-attention on the CLIP features. 
We show that the proposed video representation outperforms all other video representations on both val and test sets across all generation metrics.

\begin{table}[ht]
\scriptsize
\centering
\resizebox{\linewidth}{!}{
\begin{tabular}{l|c|c|c|c|c|c|c|c}
\hline
& \textbf{Feature} & \begin{tabular}[c]{@{}c@{}} \textbf{Video} \\ \textbf{Enc} \end{tabular}& \textbf{Q} & \textbf{R-L} &  \textbf{B-4} &  \textbf{M} &  \textbf{C} &  \textbf{BERT} \\ \hline
\multirow{5}{*}{Val}&
EgoVLP & None & No & 31.46 & 25.77 & 31.67 & 10.79 & 25.66 \\
& CLIP  & None & No & 31.60 & 25.79 & 31.91 & 10.94 & 26.34 \\
& CLIP & SA TF & No & 31.51 & 25.39 & 31.60 & 10.85 & 24.95 \\
& CLIP & SA TF & Yes & 31.54 & 25.71 & 31.75 & 10.61 & 26.20 \\
& CLIP & XA TF & Yes & \textbf{31.85} & \textbf{26.35} & \textbf{31.98} & \textbf{11.83} & \textbf{26.83} \\
\hline
\end{tabular}
}
\caption{Comparing different methods to learn the video representation from the visual features and the video encoders. SA TF - Self Attention Transformer. XA TF - Cross Attention Transformer. Video representation learned over the CLIP features via cross-attention mechanism outperforms all other methods.}
\label{tab:vis-encoder-ablations}
\end{table}

\begin{figure}
\begin{subfigure}[b]{0.49\linewidth}
\includegraphics[width=\linewidth]{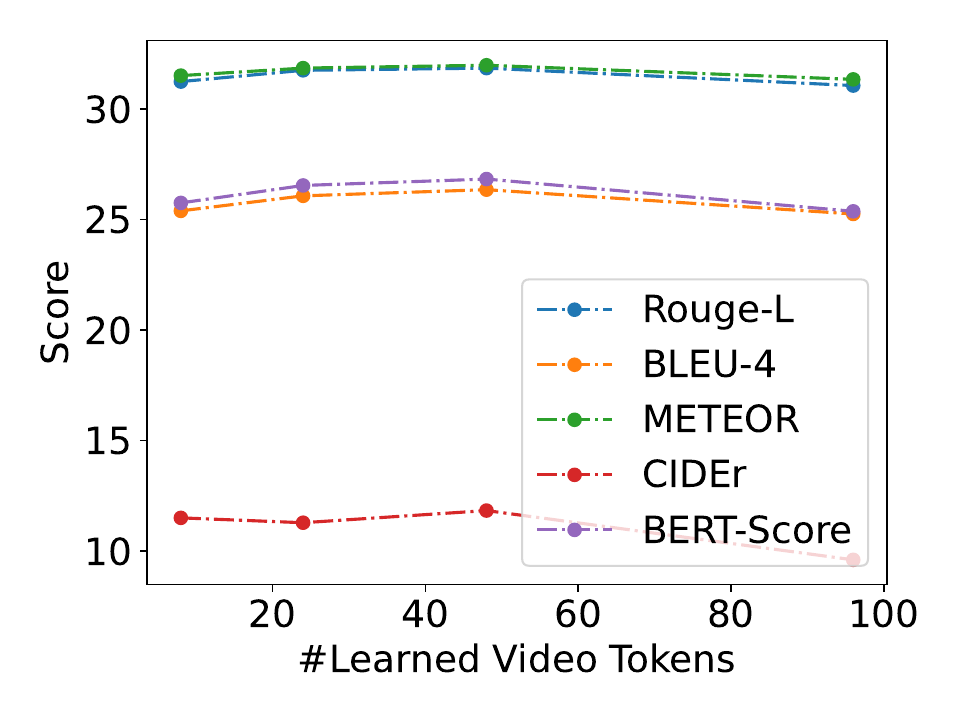}
\label{fig:lea_que}
\end{subfigure}%
\begin{subfigure}[b]{0.49\linewidth}
\includegraphics[width=\linewidth]{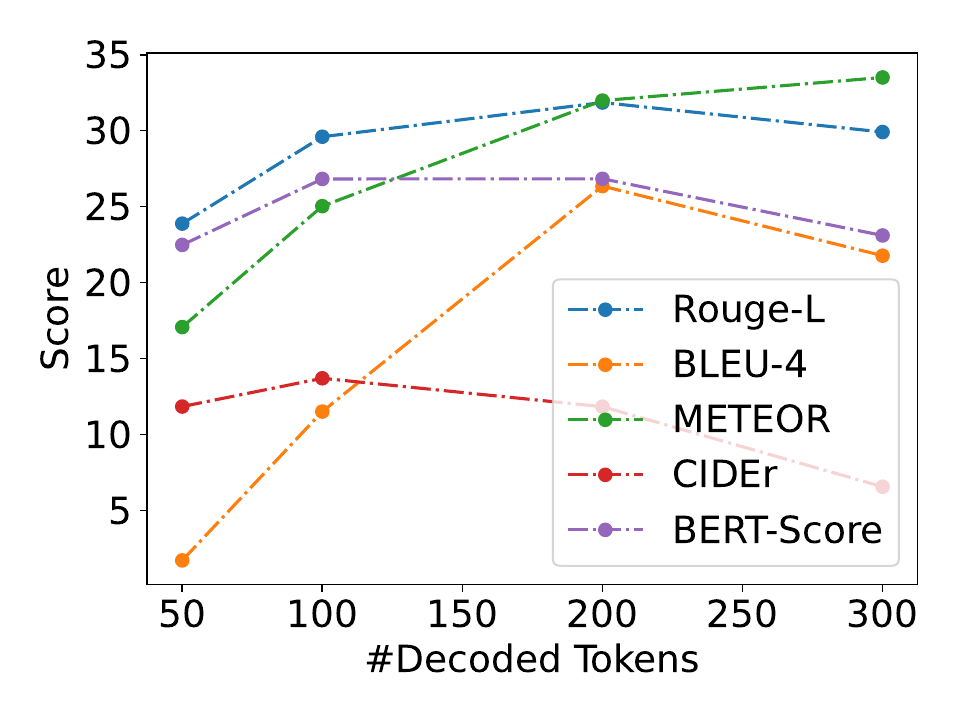}
\label{fig:dec_tok}
\end{subfigure}
\caption{(Left): Effect of number of learned video tokens on different generation metrics. Performance peaks at 48 learned video tokens. (Right): Effect of number of decoded tokens on different generation metrics. Performance peaks around 200 tokens.}
\label{fig:lea_que_dec_tok}
\end{figure}

\noindent \textbf{Learned Video Tokens.} We study the effect of varying the number of the learned video tokens on the future narration generation task in~\cref{fig:lea_que_dec_tok}. For the ViNa-GPT2 model, as we increase the number of the learned video tokens, we find that the performance peaks at 48 tokens and drops as we go further. This shows that at least the same number of tokens are needed as frames (48 seconds sampled at 1 fps) to capture the events in the observed videos.


\noindent \textbf{Decoded Tokens.} We examine how changing the number of decoded tokens influences the future narration generation performance in \cref{fig:lea_que_dec_tok}. In our ViNa-GPT2 model, for generating the future narration tokens at inference, we use standard greedy search and set nucleus sampling~\cite{holtzman2019curious} with p=0.95 and return K=50 candidate outputs. As we increase the number of tokens generated from 50 to 300, we find that model performance peaks around 200 tokens. This roughly corresponds to the median length of 20 ground truth narrations in the dataset.

\subsubsection{Role of Observed Videos and Narrations} 
We investigate the importance of the observed video and the observed narrations in generating future narrations. First, we train a GPT-2 decoder to generate future narrations conditioned only on the observed narrations. Second, we train ViNa-GPT2 to predict future narrations only based on video tokens.

\begin{table}[ht]
\scriptsize
\resizebox{\linewidth}{!}{
\begin{tabular}{c|c|c|c|c|c|c|c}
\hline
& \textbf{Obs Vid} & \textbf{Obs Nar} & \textbf{R-L} & \textbf{B-4} & \textbf{M} & \textbf{C} &  \textbf{BERT} \\
\hline
\multirow{3}{*}{Val} & No & Yes & 31.04 & 25.14 & 31.22 & 10.20 & 23.89 \\  
& Yes & No & 21.87 & 8.68 & 21.97 & 1.64 & 11.14 \\ 
& Yes & Yes & \textbf{31.85} & \textbf{26.35} & \textbf{31.98} & \textbf{11.83} & \textbf{26.83} \\
\hline
\end{tabular}
}
\caption{Role of observed video and narrations. Combining videos and narrations produces better results across all future generation metrics.}
\label{tab:vid_nar_contribution}
\end{table}

As shown in \cref{tab:vid_nar_contribution}, observed narrations can generate future narrations better than observed videos. We attribute this to the large domain gap between video and future narrations compared to observed and future narrations. Combining videos and narrations improves narration generation as they encode complimentary information.  

\subsubsection{Using Predicted Observed Narrations}
\label{sec:predicted_narr}
We seek to explore how would the visual-language models perform in the absence observed ground-truth narrations. 
Such a setup is useful for generating future narrations for out-of-domain videos where ground-truth narrations are not available.
As there are no existing off-the-shelf models to generate future narrations directly, we first predict observed narrations from the video input using an existing narration generation model - LAVILA~\cite{zhao2023learning}. We then provide the observed video and predicted narrations as input to ViNa.

\begin{table}[ht]
\scriptsize
\centering
\begin{tabular}{l|l|c|c|c|c|c} 
\hline
& \textbf{Model} & \textbf{R-L} &  \textbf{B-4} &  \textbf{M} &  \textbf{C} &  \textbf{BERT} \\ \hline
\multirow{4}{*}{Val} & ViNa-Llama3-8b & 24.78 & 12.54 & 25.70 & 3.78 & 13.99 \\
& ViNa-Llama-13b & 25.48 & 11.89 & 24.28 & 4.41 & 17.46  \\
& ViNa-T5 & 25.52 & 8.47  & 21.82 & \textbf{6.23} & \textbf{22.09}  \\
& ViNa-GPT2 & \textbf{25.52} & \textbf{13.51}  & \textbf{25.70} & 4.11 & 17.48  \\

\hline

\cline{2-7}

\hline
\end{tabular}

\caption{Comparing future narration generation performance of ViNa on predicted narrations (LAVILA). Compared to ground-truth narrations, there is a drop in performance across all models. ViNa-T5 and ViNa-GPT2 outperform across metrics.}
\label{tab:arch-design-pred}
\end{table}

\cref{tab:arch-design-pred} compares the performance of ViNa using observed video and LAVILA predicted narrations. We see a performance drop across all metrics. The difference in the quality of the narrations directly impacts future narration generation performance of all models. Similar to the trend observed in ground-truth narrations in \cref{tab:arch-design-gt}, ViNa-T5 and ViNa-GPT2 outperform other models on LAVILA generated narrations. 

\subsection{Qualitative Results}
\label{sec:qualitative_results}
\begin{figure*}
    \centering
    \includegraphics[width=0.9\linewidth]{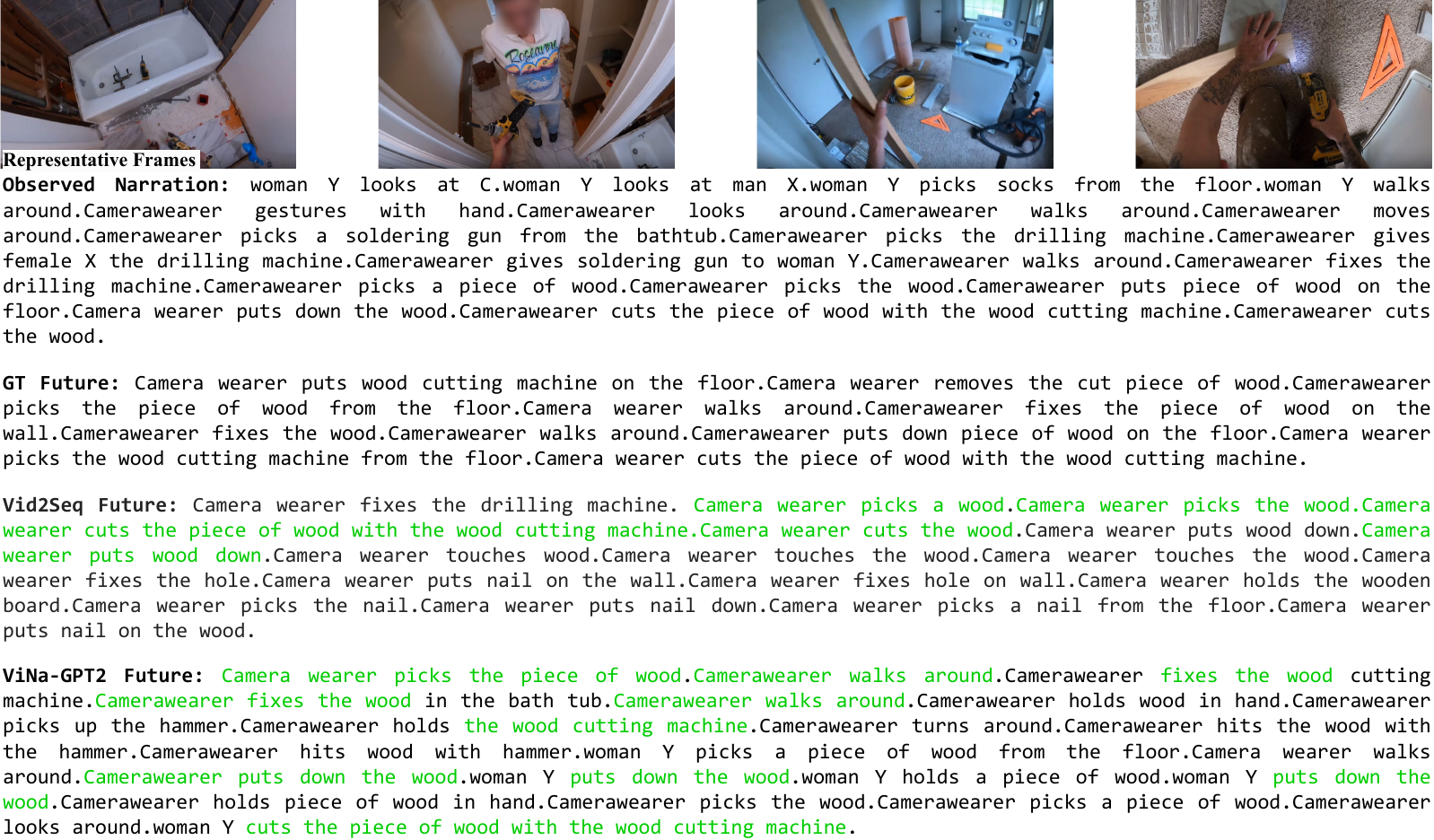}
    \caption{ViNa-GPT2 maintains continuity of observed narrations and matches the future narrations better than Vid2Seq. \textcolor{green}{Green} shows matched narrations with ground truth.}
    \label{fig:qualitative_example}
\end{figure*}

\begin{figure*}
\centering
\includegraphics[width=0.9\textwidth]{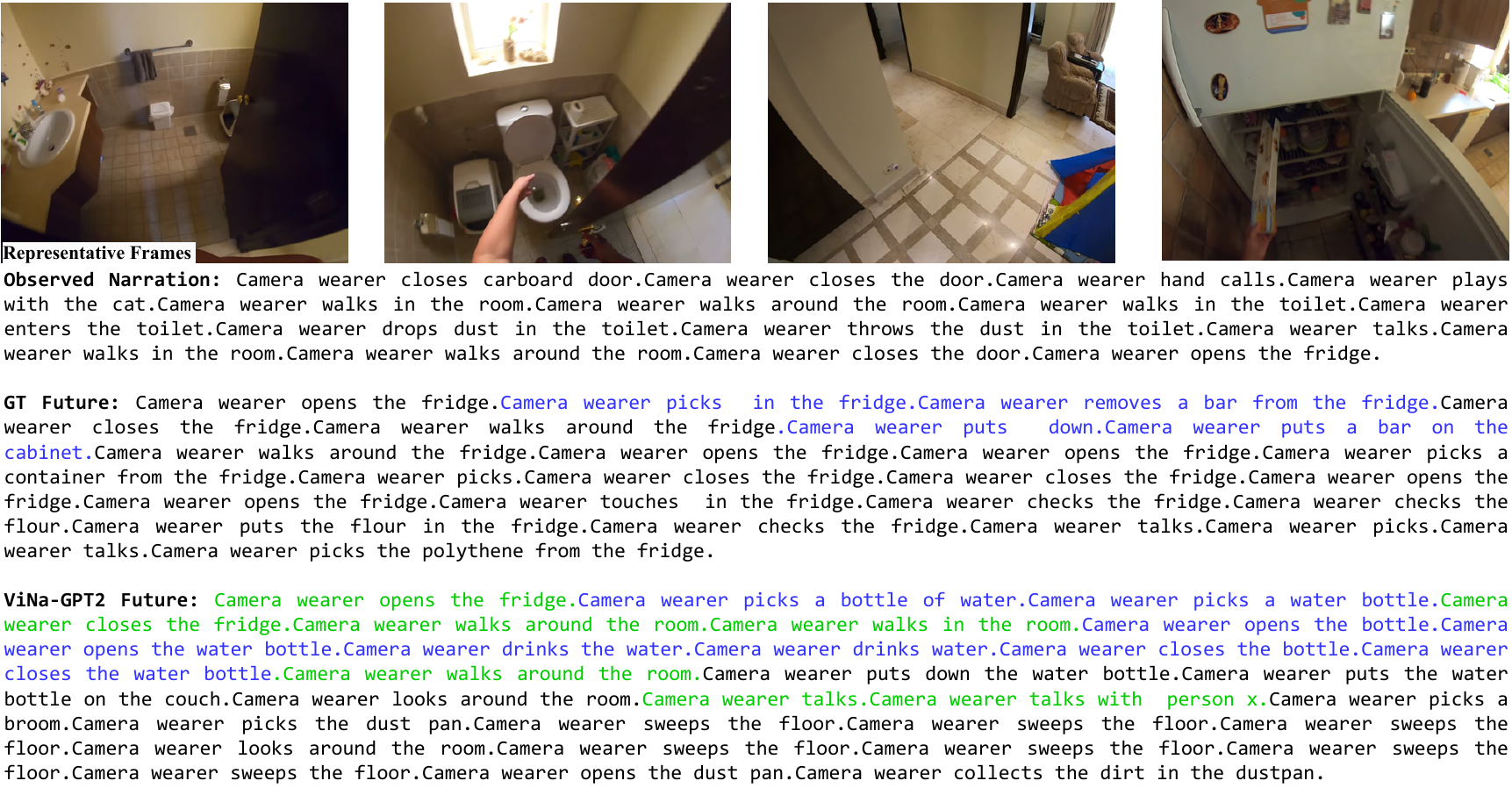}
\captionof{figure}{Future narrations generated by ViNa-GPT2 are consistent with the visual context provided by the observed video. 
\textcolor{green}{Green} shows generated narrations that match with GT. \textcolor {blue}{Blue} indicates alternate future narrations not present in GT but plausible given the observed video.}
\label{fig:diverse-example-2}
\end{figure*}

We provide an example of future narrations generated by ViNa-GPT2 and Vid2Seq in ~\cref{fig:qualitative_example}. Both Vid2Seq and ViNa-GPT2 generate diverse narrations but ViNa-GPT2 narrations cover more of the ground truth and is also closer to ground truth. 
We provide another example in \cref{fig:diverse-example-2} to show how the generated future narrations from ViNa-GPT2 encompass a series of events that seem to logically follow the observed events even when this behaviour is not seen in the ground truth future narrations. Considering the last observation of the visual context depicts a person opening a refrigerator, the projected future by ViNa-GPT2 entails narrations such as \textit{picking a water bottle from the fridge, opening it, drinking from it, and then closing the bottle}. Also, the sequence of events \textit{``Camera wearer picks a broom. Camera wearer picks the dust pan. Camera wearer sweeps the floor. Camera wearer opens the dust pan. Camera wearer collects the dirt in the dustpan.''} in the generated future circles back to close the loop of the observed events \textit{``Camera wearer enters the toilet. Camera wearer throws the dust in the toilet.''}
Additional examples in supplementary.

\section{Conclusion and Future Work}
We have proposed a novel task of long-term future narration generation in daily activities, aimed at providing detailed descriptions of future human actions. Our model, ViNa, outperforms existing approaches adapted for future narration generation on the largest egocentric dataset Ego4D. We have shown the significance of narration generation as a fundamental task that enhances other future prediction tasks such as long-term action anticipation, goal prediction, and next goal-step prediction. Furthermore, we have demonstrated an application of how the generated narrations can be utilized for future video clip retrieval. 
The scores of the evaluated models are low on the task of future narration forecasting which means it is quite a challenging task.
In future, we would like to investigate ways to incorporate contextual information and user preferences into the narration generation process that could lead to more personalized and accurate predictions. 

{\small
\bibliographystyle{ieee_fullname}
\bibliography{egbib}
}

\pagebreak

\twocolumn[
\begin{center}
\textbf{\Large Supplementary Material: Learning to Generate Long-term Future Narrations Describing Activities of Daily Living}
\end{center}
\vspace*{12pt}
]

\setcounter{section}{0}
\setcounter{equation}{0}
\setcounter{figure}{0}
\setcounter{table}{0}
\makeatletter
\renewcommand{\theequation}{S\arabic{equation}}
\renewcommand{\thefigure}{S\arabic{figure}}
\renewcommand{\thesection}{S\arabic{section}}
\renewcommand{\thetable}{S\arabic{table}}

\section{Expressivity of Ego4D-Narrations} 

\begin{figure}[ht]
\centering
\includegraphics[width=\linewidth]{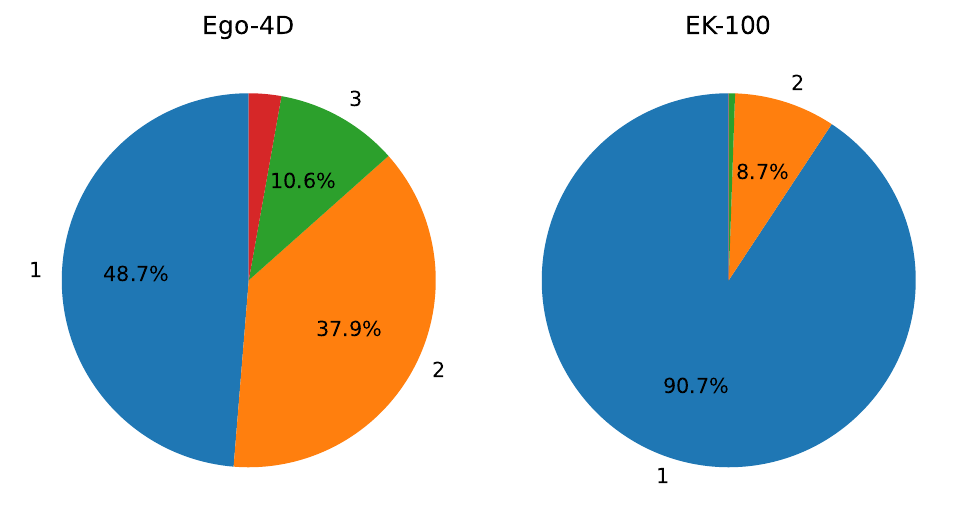}
\captionof{figure}{Distribution of narrations over relations in Ego-4D(Left) and EK-100(Right). More than 50\% of the narrations in Ego-4D, vis-a-vis 9.3\% in EK-100, encode relationships between 2 or more nouns.}
\label{fig:relation_disb}
\end{figure}

We analyze the complexity of relations encoded by the narrations using Stanford's Scene Graph Parser~\cite{wu2019unified} based on dependency parsing. Out of the 2,796,021 narrations from Ego4D~\cite{grauman2022ego4d}(\textit{FHO} and \textit{EM} subsets), about half (48.71\%) encode relations which involve only a single noun similar to actions. As shown in \cref{fig:relation_disb}, 51.29\% of narrations encode relationships involving 2 or more nouns (person, object, place, etc.). On the contrary, in EK-100~\cite{damen2022rescaling}, only 9.3\% of the narrations encode complex relationships involving 2 or more nouns.
Therefore, we propose the Future Narration Generation task to generate events involving multiple future relationships through future narrations.

\section{Experimental Details for Training and Evaluating ViNa}
In our ground truth narrations, we replace the term \textbf{C} with  \textbf{Camera wearer} to aid the generation process of the language decoder (GPT2, T5, Llama-13b, Llama3-8b, etc.). 
While evaluating the ViNa models across generation metrics, we remove the term \textbf{camera wearer} from the generated future narrations and the ground truth future narrations. This ensures the scores are not influenced by \textit{camera wearer} which occurs in almost every narration in Ego-4D~\cite{grauman2022ego4d}.

\section{Learning the Separator Tokens}

When creating the input token sequence for ViNa-GPT2 containing the learned video and observed narration tokens, we add fixed separator tokens $\texttt{<vid>}$ and $\texttt{<obs>}$ representing the start and end of the video and the observed narration respectively. We investigate the performance of ViNa-GPT2 when these separator tokens are instead learned in~\cref{tab:sep-tokens}. We find empirically that fixing the separator tokens outperforms treating them as learnable tokens.

\begin{table}[ht]
\centering
\scriptsize
\begin{tabular}{l|c|c|c|c|c|c}
\hline
& \textbf{Separator Tokens} & \textbf{R-L} &  \textbf{B-4} &  \textbf{M} &  \textbf{C} &  \textbf{BERT} \\ \hline
\multirow{2}{*}{Val} & Learned & 30.08 & 23.32 & 30.39 & 10.09 & 23.84 \\
& Fixed & \textbf{31.85} & \textbf{26.35} & \textbf{31.98} & \textbf{11.83} & \textbf{26.83} \\
\hline
\end{tabular}
\caption{Effect of learned vs fixed separator tokens on the ViNa-GPT2 model. Performance of the fixed tokens is superior to the learned separator tokens.}
\label{tab:sep-tokens}
\end{table}

\section{Temporal Resolution of the Video Frames}
We study the effect of changing the temporal resolution (frame rate) of the observed video in ~\cref{tab:frame-rate}. We start with a frame rate of 0.5 fps, increase the sampling frequency to 1 fps and finally sample only key-frames in the observed video. The key-frames are the full image frames in the observed video obtained using the \textit{decord video reader} library. The number of learned video tokens across all three settings is fixed at 48. We report that ViNa-GPT2 performs comparably even at 0.5 fps. Since the number of sampled key-frames can vary widely depending on the video content, we use a uniform sampling rate of 1 fps in all our experiments which is easily reproducible and produces the best results. 
\begin{table}[ht]
\centering
\scriptsize
\begin{tabular}{l|c|c|c|c|c|c}
\hline
& \textbf{Sampling Strategy (fps)} & \textbf{R-L} &  \textbf{B-4} &  \textbf{M} &  \textbf{C} &  \textbf{BERT} \\ \hline

\multirow{3}{*}{Val} & 0.5 & 31.29 & 25.27 & 31.41 & 10.80 & 25.49 \\
& Key-Frames & 31.27 & 25.29 & 31.34 & 10.69 & 25.55 \\
& 1 & \textbf{31.85} & \textbf{26.35} & \textbf{31.98} & \textbf{11.83} & \textbf{26.83} \\

\hline
\end{tabular}
\caption{Comparing performance of the ViNa-GPT2 model by varying the temporal resolution of the observed video. Observed video frames sampled at a uniform frame rate of 1 fps outperforms other sampling rates.}
\label{tab:frame-rate}
\end{table}

\section{Qualitative Results}
We now provide more qualitative samples of the future narrations generated by the ViNa-GPT2 model.

\begin{figure*}
\centering
\includegraphics[width=\textwidth]{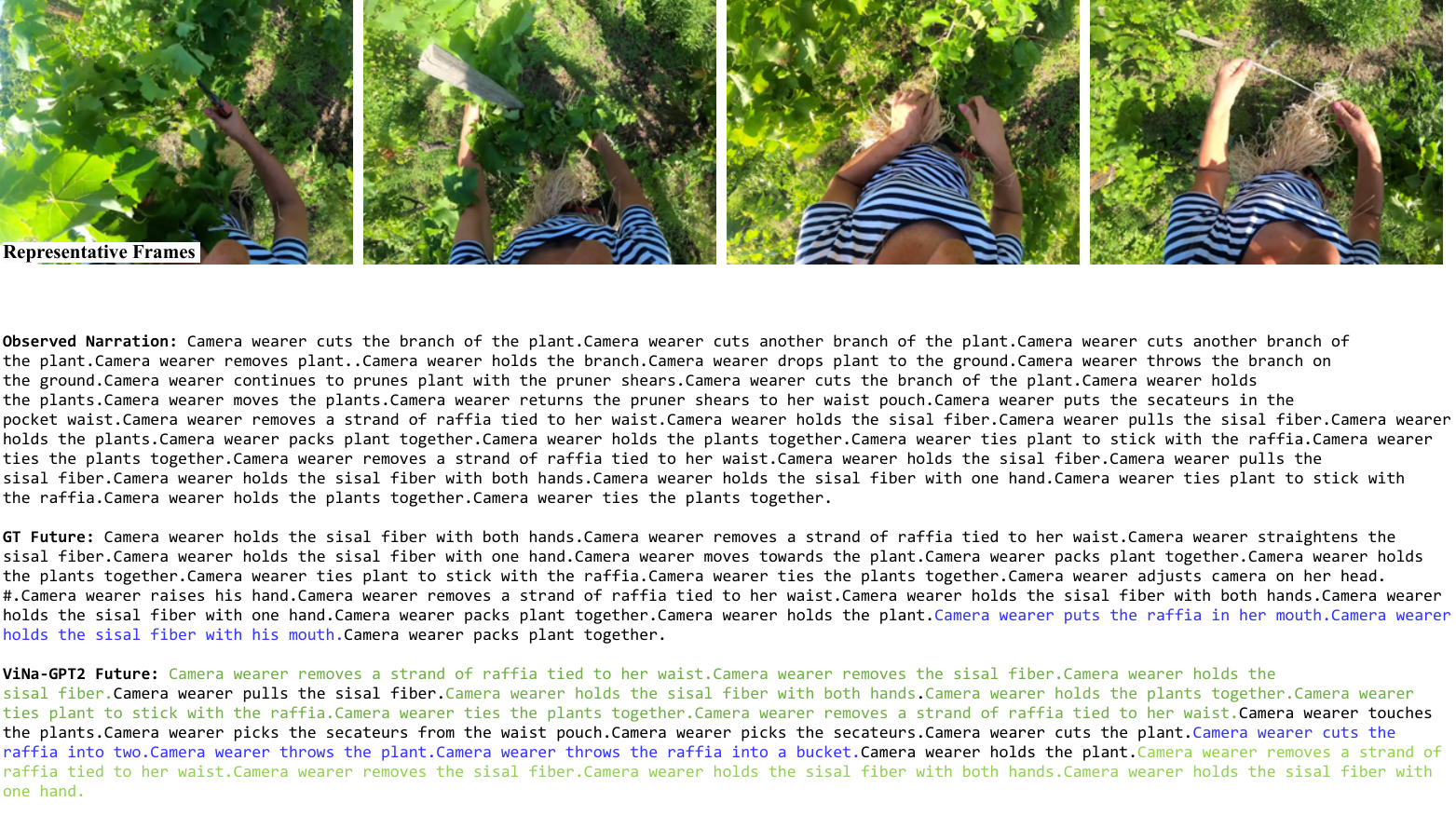}
\caption{ViNa-GPT2 maintains continuity and also generates plausible alternate future narrations \textcolor{blue}{``Camera wearer throws the raffia into a bucket''} compared to GT Future  \textcolor{blue}{``Camera wearer puts the raffia in her mouth''}. \textcolor{green}{Green} shows matched narrations with ground truth. \textcolor{blue}{Blue} indicates alternate future narrations not present in GT.}
\label{fig:simple-example-1}
\end{figure*}

\begin{figure*}
\centering
\includegraphics[width=\textwidth]{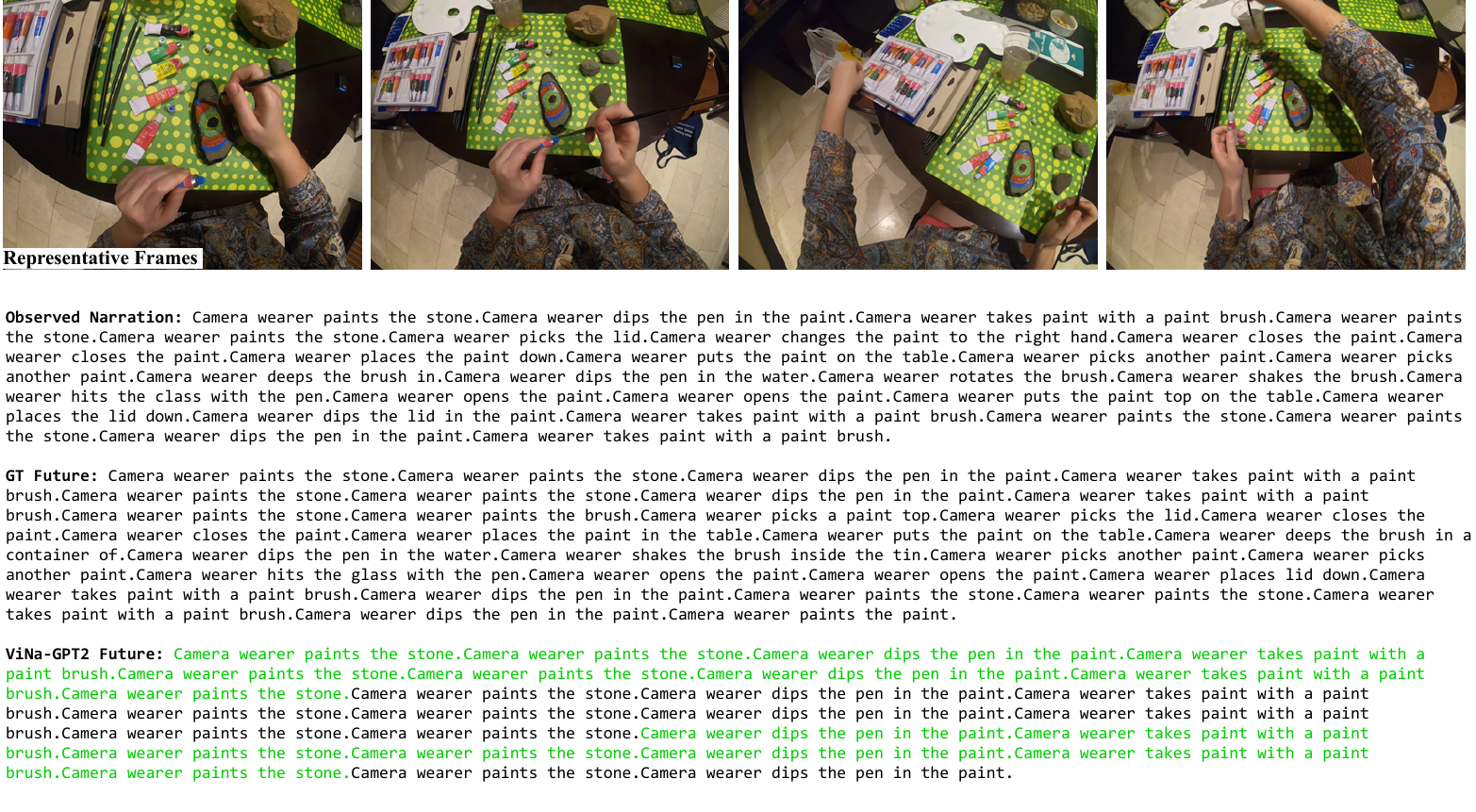}
\captionof{figure}{For long repetitive sequence of events, ViNa-GPT2 generates the future narrations that follow the sequence. \textcolor{green}{Green} shows matched narrations with ground truth.}
\label{fig:simple-example-2}
\end{figure*}

\begin{figure*}
\centering
\includegraphics[width=\textwidth]{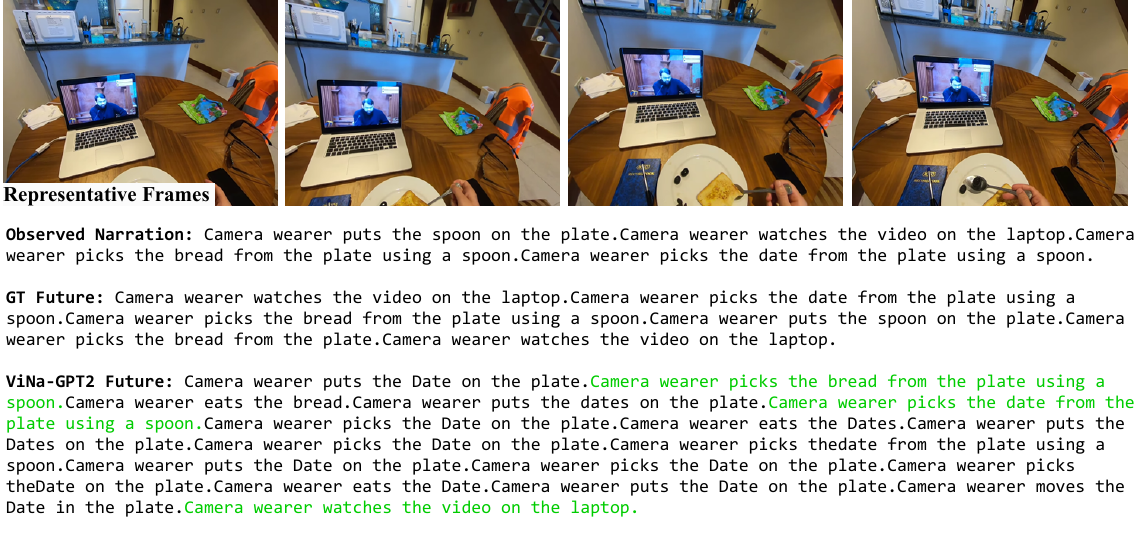}
\captionof{figure}{Future narrations generated by ViNa-GPT2 are consistent even when the number of observed narrations is fewer than normal for the observation period of 48 seconds. \textcolor{green}{Green} shows matched narrations with ground truth.}
\label{fig:simple-example-3}
\end{figure*}

In \cref{fig:simple-example-1}, \cref{fig:simple-example-2} and \cref{fig:simple-example-3}, we observe that ViNa-GPT2 is able to  faithfully continue the observed events in future as well as generate new plausible events which don't occur in the observed narrations.

\begin{figure*}
\centering
\includegraphics[width=\textwidth]{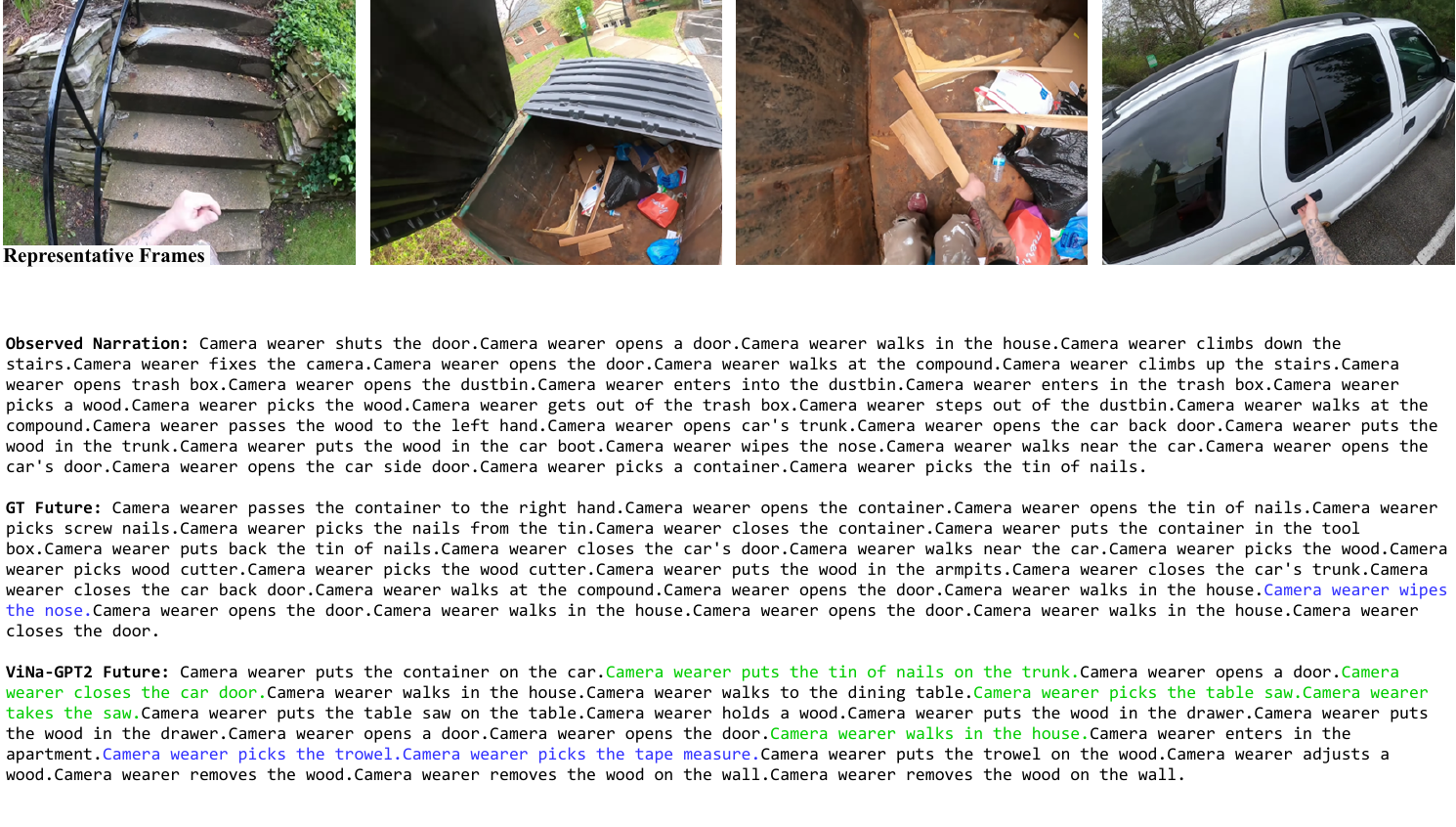}
\captionof{figure}{ViNa-GPT2 adapts to dynamic scenarios where diverse events unfold within the same observation period of 48 seconds. The generated future narration \textcolor{green}{``Camera wearer picks the table saw''} is close to the GT future ``Camera wearer picks wood cutter''. An interesting observation is that even when \textit{wood cutter} is not present in the observed narrations, ViNa-GPT2 is able to anticipate the object \textit{table saw}. 
\textcolor{green}{Green} shows matched narrations with ground truth. \textcolor{blue}{Blue} indicates alternate future narrations not present in GT.}
\label{fig:diverse-example-1}
\end{figure*}

\begin{figure*}
\centering
\includegraphics[width=\textwidth]{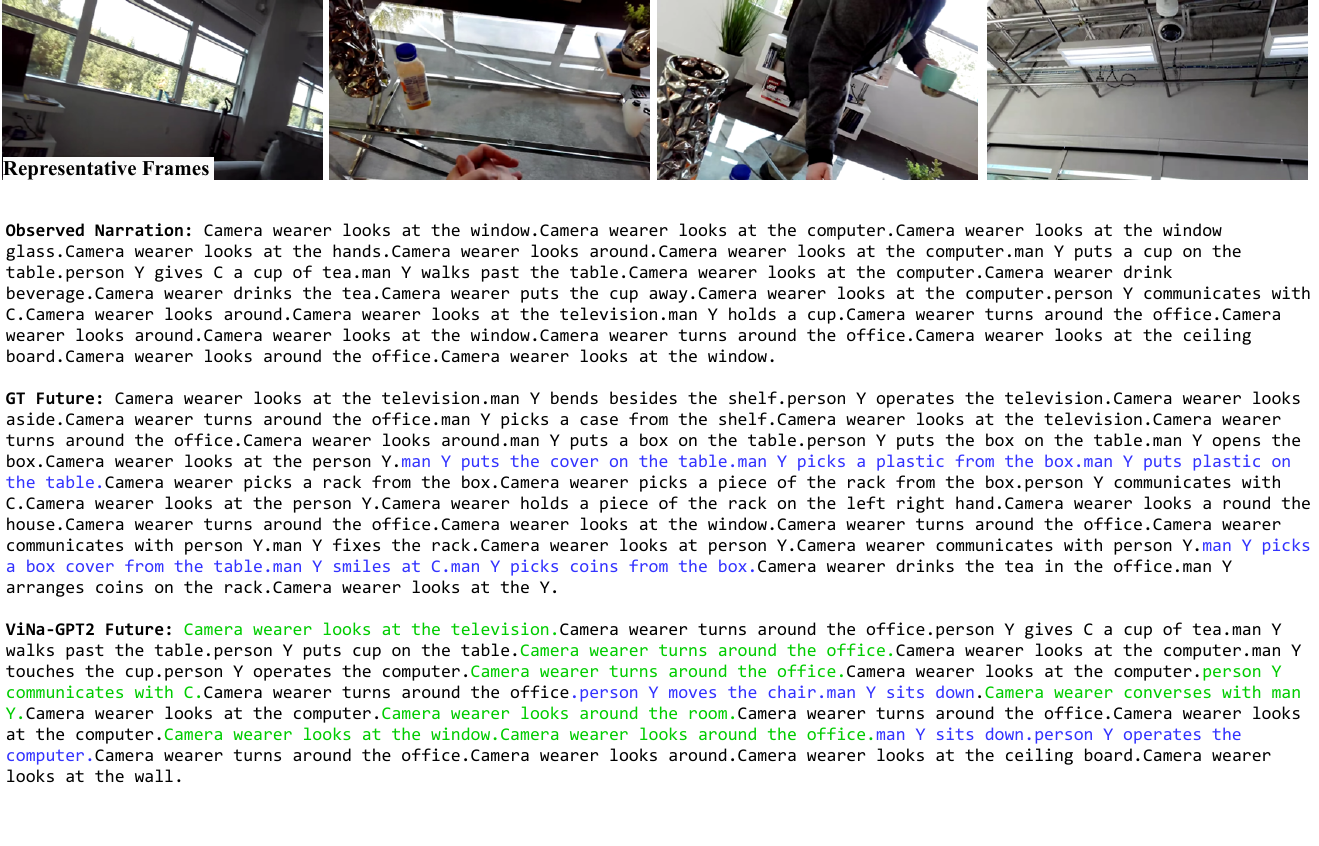}
\captionof{figure}{ViNa-GPT2 generated future narrations follow  GT future narrations closely. The generated narrations involving \textcolor{blue}{man Y} contain plausible alternate future narrations \textcolor{blue}{``person Y moves the chair.man Y sits down.''} compared to the GT future narrations \textcolor{blue}{``man Y puts the cover on the table.man Y picks a plastic from the box.man Y puts plastic on the table.''} \textcolor{green}{Green} shows matched narrations with ground truth. \textcolor{blue}{Blue} indicates alternate future narrations not present in GT.}
\label{fig:diverse-example-3}
\end{figure*}

From \cref{fig:diverse-example-1} and \cref{fig:diverse-example-3}, we see that ViNa-GPT2 is able to adapt to rapidly evolving scenarios and generate future narrations consistent with the observed visual context. These situations involve several diverse events occurring within the same fixed observed duration. 

\end{document}